\documentclass[runningheads]{llncs}

\usepackage{eccv}

\usepackage{eccvabbrv}

\usepackage{graphicx}
\usepackage{booktabs}
\usepackage{makecell}
\usepackage{multirow}
\usepackage{caption}

\usepackage{amsmath}
\DeclareMathOperator*{\argmin}{argmin}

\usepackage[accsupp]{axessibility}  

\usepackage{hyperref}

\usepackage{orcidlink}

\begin{document}

\title{From Perspective to Fisheye Depth Estimation and Open-Vocabulary Segmentation} 


\author{
    Rit Gangopadhyay \orcidlink{0000-0001-5685-5014}\index{Gangopadhyay, Rit}
    \and
    Alex Wong \orcidlink{0000-0002-3157-6016}\index{Wong, Alex}
}

\authorrunning{R. Gangopadhyay and A. Wong}
\titlerunning{Distortion Extenders}


\institute{
    Yale Vision Laboratory, Yale University, New Haven, CT 06520, USA \\
    \email{\{rit.gangopadhyay,alex.wong\}@yale.edu}
}

\maketitle

\begin{abstract}
Vision foundation models are capable of generalizing across 3-dimensional (3D) scenes with high-fidelity estimates; their empirical success can be attributed to training on large-scale datasets of perspective images. However, when transferred to wide field-of-view (FoV) images, such as those captured by fisheye cameras, they return erroneous outputs due to a covariate shift stemming from the radial distortion on the image pixels. We propose a method to generalize vision foundation models to fisheye cameras. The crux of our method lies in a set of learnable parameters, termed Distortion Extenders (DEX), that model the fisheye distortion coefficients and the distributional shift between fisheye and perspective images encoded in the latent space. By minimizing a self-supervised alignment loss, DEX transforms the latent embeddings of fisheye images to resemble those of perspective images to recover high-fidelity estimates. DEX is architecture- and task-agnostic: We demonstrate DEX on monocular depth estimation and open-vocabulary segmentation for convolution- and Transformer-based architectures, where we consistently improve over baselines across indoor and outdoor fisheye datasets. As a byproduct, the activations of DEX can also be decoded to distortion coefficients to support camera calibration. Code available at: \url{https://github.com/Suchisrit/DEX}.
\end{abstract}

\section{Introduction}

\begin{figure}[th]
    \centering
    \includegraphics[width=0.9\linewidth]{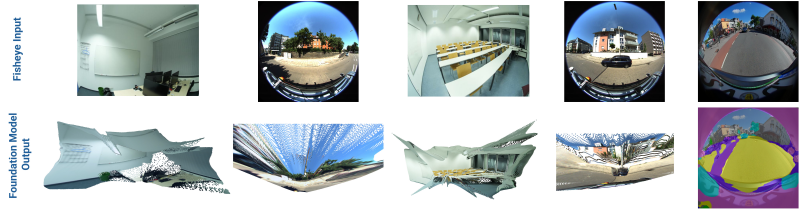}
    \vspace{-7mm}
    \caption{3D reconstructions and open-vocabulary segmentation of fisheye imagery produced by foundational models, UniDepthV2 \cite{piccinelli2025unidepthv2} and LSeg \cite{li2022language}. The distortion in the input leads to erroneous and distorted reconstructions and segmentations.}
    \label{fig:mde_baseline}
\end{figure}

Vision foundation models are typically trained on tens to hundreds of millions of images, enabling them to generalize across a wide range of 3-dimensional (3D) scenes and across tasks, spanning from recognition to 3D reconstruction. These models leverage the large volumes of internet images available, which are predominantly captured by standard perspective cameras of 3D scenes. The relationship of 3D points and their projection onto a 2D image plane can be described by the projective geometry of the camera, where the perspective cameras capturing the majority of the large-scale training dataset are well-approximated by a simple pinhole projection model. When training on such datasets, biases of the camera projective geometry are ``baked into'' these foundation models, which are often deployed onto mixed camera systems to enable spatial applications, from extended realities to robotic manipulation and autonomous navigation. 

Yet, within these mixed camera systems are wide field-of-view (FoV) cameras necessary for environmental coverage (e.g., fisheye and other wide angle cameras). Their projective geometry differs from that of perspective cameras, in that 3D points follow a highly non-linear projection model that induces strong radial distortions. As the FoV increases, objects that are projected near the image borders become increasingly warped or distorted, deviating from the assumptions of projective geometry under which foundational models are trained. When these models are applied to fisheye images, the mismatch in projection induces a systematic distribution shift in local pixel arrangements, and in turn, the encoded features --  leading to distorted 3D reconstructions and degraded recognition performance, as seen in Fig.~\ref{fig:mde_baseline}.

A natural response is to “undo” this image distortion; hence, one line of work rectifies input images using known camera intrinsics, or re‑projects them into a canonical representation such as equirectangular, tangent, spherical, or cube‑map projections before processing \cite{li2022mode,eder2020tangent,liu2024estimating,lichy2024fova,guo2025depth,yan2022spheredepth}. While this retains compatibility with models (pre)trained on perspective images, it introduces multiple resampling steps, calibration dependencies, and characteristic artifacts (e.g., stretching near poles, seams between faces, aliasing, and gaps between patches). Another approach is to train dedicated models (e.g., depth estimators or segmentation networks) for fisheye cameras \cite{lee2023slabins,zhao2025fisheyedepth,blott2018semantic}, which often suffer in generalization across 3D scenes. This is partly due to the difficulty in assembling large‑scale fisheye training datasets, as there exist fewer available fisheye images than perspective images. While one may finetune publicly available (foundational) models for specific cameras to improve generalization across scenes, this risks parameter drift and may degrade performance on previously learned cameras \cite{gangopadhyay2025extending}. An alternative is to supplement the shortage via mixed‑camera training \cite{piccinelli2025unik3d}, yet this trades generalizability for fidelity across cameras. 

More importantly, the main school of thought aims to introduce inductive biases tailored to each camera model, which entails specialized designs and thus limits the end user's flexibility in choosing architectures. Separately, the design of these methods is also task-specific, whether reconstruction or recognition, leading to disparate approaches across vision tasks. Given that a pretrained model can already faithfully infer properties of the 3D scene from perspective images, erroneous estimates on fisheye images stem from a covariate shift in the values of latent embeddings due to local operations on image pixels displaced and rearranged by distortion. As such, if one can modulate their values to align with those that produced high-fidelity estimates in perspective images, then we hypothesize that one can extend the capabilities of models pretrained on images captured by perspective cameras to those of fisheye cameras. To ensure high utility, we consider an architecture- and task-agnostic approach, such that we can apply our method to both convolution-based and Transformer-based architectures for reconstruction and recognition tasks. 

To do so, we begin by forgoing explicit modeling of camera projective geometry in the input and latent space of the model and build on the insight that the values of embeddings, rather than coordinates, can be adjusted to recover high-fidelity estimates. To this end, we freeze the backbone to preserve its performance on perspective images and propose a set of lightweight, learnable modules that can be inserted into each block of a frozen backbone to modulate feature embeddings of fisheye images. Rather than re-projecting pixels to a canonical image plane, these parameters implicitly model the distortion in fisheye images and operate directly in latent space to map fisheye embeddings toward the distribution of perspective image embeddings. Hence, our method bypasses the need for rectification or camera-specific inductive biases and the cost to retrain the backbone to extend perspective vision models to fisheye cameras. We term our method \textit{Distortion Extenders}, or DEX for short. 

We train DEX in a self-supervised manner using only calibrated perspective images. Given a perspective image, we synthetically distort it to obtain a fisheye counterpart. We pass the perspective image through a frozen vision model and treat its output (or intermediate features) as a reference. The fisheye image is then processed by the same backbone but augmented with DEX, and an inverse geometric transformation aligns the fisheye output back to the perspective reference frame. We minimize an alignment loss that drives the Extenders to reconcile the two estimates. This procedure requires neither ground truth nor real fisheye training data, and can be applied to both convolutional and Transformer architectures and across reconstruction (e.g., monocular depth estimation) and recognition tasks (e.g., open-vocabulary segmentation). At test time, fisheye images are fed directly into the augmented model -- no calibration, re-projection, or knowledge of additional inference-time geometry is required.

We demonstrate our method on two vision tasks: monocular depth estimation and open-vocabulary segmentation. For monocular depth estimation, we further change the output parameterization from Cartesian depth \(z\) to spherical range \(R\), leveraging the structure of fisheye projection models to more naturally express distances in high-FoV settings and to counteract the radial attenuation observed in models trained on perspective images. For open-vocabulary segmentation, we formulate our objective in the latent space to better take advantage of the affinities between an image and its associated text.

\textbf{Our Contributions} are as follows: (1) We introduce \emph{Distortion Extenders} (DEX), a set of lightweight, feature-space modulators that generalizes pretrained models from perspective to fisheye imagery by transforming fisheye embeddings to align with those of perspective images. (2) We provide analysis showing that DEX indeed does model fisheye distortion coefficients and serves as latent distribution alignment. (3) DEX is architecture-agnostic and applicable to both convolution- and Transformer-based architectures. (4) DEX is also task-agnostic and enables strong zero-shot performance on real fisheye datasets for both monocular depth estimation (ScanNet++, KITTI-360) and open-vocabulary segmentation (WoodScape). As a byproduct, activations of DEX can be decoded to distortion coefficients for calibration purposes. 

\section{Related Works}

\textbf{Vision Foundation Models.} Recently, foundation models (pretrained models designed to provide strong initializations for a variety of tasks) have seen significant growth in the vision domain. These models leverage large datasets through both self-supervised and supervised learning. Their flexibility enables them to adapt to and perform a wide range of tasks. CLIP \cite{radford2021learning} demonstrates broad task applicability by using contrastive learning on a large vision-language dataset and has been a core component in open-vocabulary segmentation \cite{li2022language,xie2024sed}. DINO \cite{caron2021emerging, oquab2023dinov2} uses self-supervision to extract effective features and has been used as an encoding module in depth estimators \cite{yang2024depthanything,yang2024depthanythingv2,piccinelli2024unidepth}. However, despite being trained on extensive datasets, they experience performance degradation when used with data captured by fisheye cameras, where errors propagate to downstream models.

\textbf{Transformation-based Solutions.} To enable depth estimation on unrectified images, methods typically rely on alternate representations such as equi-rectangular projection (ERP) \cite{li2022mode,wang2020360sd}, tangent projection \cite{eder2020tangent,li2022omnifusion}, spherical projection \cite{liu2024estimating,yan2022spheredepth} or cube mapping \cite{lichy2024fova,wang2020bifuse}. However, each of these representations has drawbacks: aside from resampling artifacts, ERP suffers from vertical stretching distortion and angular compression towards the poles, tangent projection exhibits radial stretching and nonuniform area scaling towards the edges, and cube maps have discontinuous transitions between each face of the cube. These artifacts can introduce spurious errors when processed by standard neural network operations, such as convolutions, tokenization, and patch embedding. Spherical projections further require specialized network operations, which increases implementation complexity. Additionally, these projections introduce latency during inference and require accurate camera calibration at test time, which can be sensitive to small physical changes in the system. 

\textbf{Depth Estimation for Specific Cameras.} 
Hence, to avoid projecting input images to a canonical representation, another line of work seeks to train depth estimators tailored to specific camera models \cite{fei2019geo,ezhov2024all,gui2022efficient,guo2026vista3d,lee2023slabins,liu2022monitored,park2026orcas,peer2002panoramic,rim2025protodepth,rim2025radar,shen2022panoformer,singh2023depth,upadhyay2023enhancing,yang2019dense,wong2019bilateral,wong2020unsupervised,wong2021adaptive,wong2021learning,wong2021unsupervised,zhao2025fisheyedepth}, e.g., perspective, wide-angle, fisheye, omnidirectional. Yet, it is challenging to acquire training datasets for wide FoV cameras compared to publicly available perspective camera datasets. As such, it is difficult to train foundation models for wide FoV cameras. Nonetheless, one can finetune existing large-scale pretrained models, e.g., foundational depth estimators, for specific cameras \cite{berenguel2023convolution,de2018eliminating,wu2025depthfisheye}. While finetuning can improve performance for a target camera, it also introduces the risk of parameter drift \cite{chung2025eta,park2024test}, where the resulting models may lose their generalizability across 3D scenes and previously learned cameras. It also produces camera-specific models, so deployment platforms (e.g., autonomous vehicles, XR, robotic systems) with mixed camera systems must maintain multiple depth estimators, which increases operational complexity.

\textbf{Generalizable Depth Estimators.}
To obtain models that generalize across 3D scenes and cameras, recent work has focused on developing foundational monocular depth estimators (FMDEs) \cite{yang2024depthanything,piccinelli2024unidepth,ranftl2021vision}. Aside from training on large-scale datasets of mixed cameras, methods in this vein also incorporate the previously mentioned projection to some canonical space, either in the input or in the internal latent representation: DepthAnyCamera \cite{guo2025depth} applies equirectangular projection (ERP) to the input image before processing it through the network and backprojects the output to the original image space; FoVA-Depth \cite{lichy2024fova} instead utilizes cube mapping of the input image; Unik3D \cite{piccinelli2025unik3d} adopts a spherical internal representation and trains on 23 different datasets of mixed cameras to achieve state-of-the-art performance. However, the fidelity across all cameras is reduced. Additionally, the architectural complexity introduced by these models amounts to hundreds of millions of parameters, which imposes less flexibility on deployment platforms with specific computational requirements.

\textbf{Image Segmentation} has also benefited from large-scale pretrained networks. Classic convolution-based architectures like U-Net \cite{ronneberger2015u} and DeepLab \cite{chen2017deeplab} remain popular, while Transformer-based methods \cite{xie2021segformer} have also shown success in capturing global context. Recent developments such as the Segment Anything Model \cite{kirillov2023segment} highlight the power of large-scale self-supervision and prompt-based adaptation. Semantic segmentation \cite{chen2017deeplab,lao2024depth,lao2024sub,ronneberger2015u,wong2021small} involves recognizing the identity of segmented regions in the image, extensions of which include \emph{Open-vocabulary Semantic Segmentation}, where the possible classes for objects in the scene are unknown until inference time. Many of such methods \cite{liang2023open, lin2023clip, he2023clip, zhang2025corrclip, zhang2024exploring} rely on the CLIP vision-language model because of its strong alignment between image and language embeddings, which can be used to assign labels to regions of an image. Recent methods such as LSeg \cite{li2022language} and SED \cite{xie2024sed} encode text and language separately at input and compare their embeddings to assign regions of the image to the given set of classes. Similar to depth estimation, segmentation pipelines can suffer when encountering heavy distortions. However, most existing work to handle fisheye images can only be applied to depth estimation. We instead show how our method can be easily adapted to recognition tasks.

\textbf{Differentiation from Model Extension and Adaptation Methods.} The \textit{closest} (concurrent) work to ours is Calibration Tokens \cite{duan2026fisheye3r,gangopadhyay2025extending}, which extend FMDEs pretrained on perspective images to fisheye images by appending trainable tokens to Transformer blocks. 
Unlike them, the proposed Distortion Extenders are distinct in: (1) Our method is architecture-agnostic and can be applied to feature map-based architectures (e.g., convolutional neural networks, CNNs) as well as token-based architectures (e.g. Transformers); Calibration Tokens are limited to Transformers due to their use of tokens. (2) Calibration Tokens require manual removal of its appended tokens after each layer, which adds complexity; unlike them, our method does not require appending or discarding of any element, but automatically determines the weighting of the Extenders as a convex combination for modulation based on their affinity towards fisheye features. (3) Calibration Tokens utilize a loss specific to depth, which limits their tasks, whereas our loss is applicable to both depth estimation and segmentation. (4) A minor difference is that Calibration Tokens output depth in z-depth (Cartesian); whereas, we output in range (Euclidean) for depth estimation.

Also related are cross-attention \cite{vaswani2017attention} and LoRA \cite{hu2022lora}. Cross-attention determines an added bias by learning projections of different inputs to queries, keys, and values; whereas, LoRA projects the input directly to an added bias by learning a weight matrix. Unlike them, we learn Extender modulators, and a projection matrix to select amongst them, that allow the input features to directly determine weights for convex combinations of the Extenders. The result of which will modulate each encoder block within a backbone. Intuitively, one may analogize Extenders to ``centers'' or entries of a ``codebook'' that model distortion coefficients and the distributional shift between perspective and fisheye images.

\section{Method}

\begin{figure*}[th]
  \centering
  \includegraphics[width=0.8\textwidth]{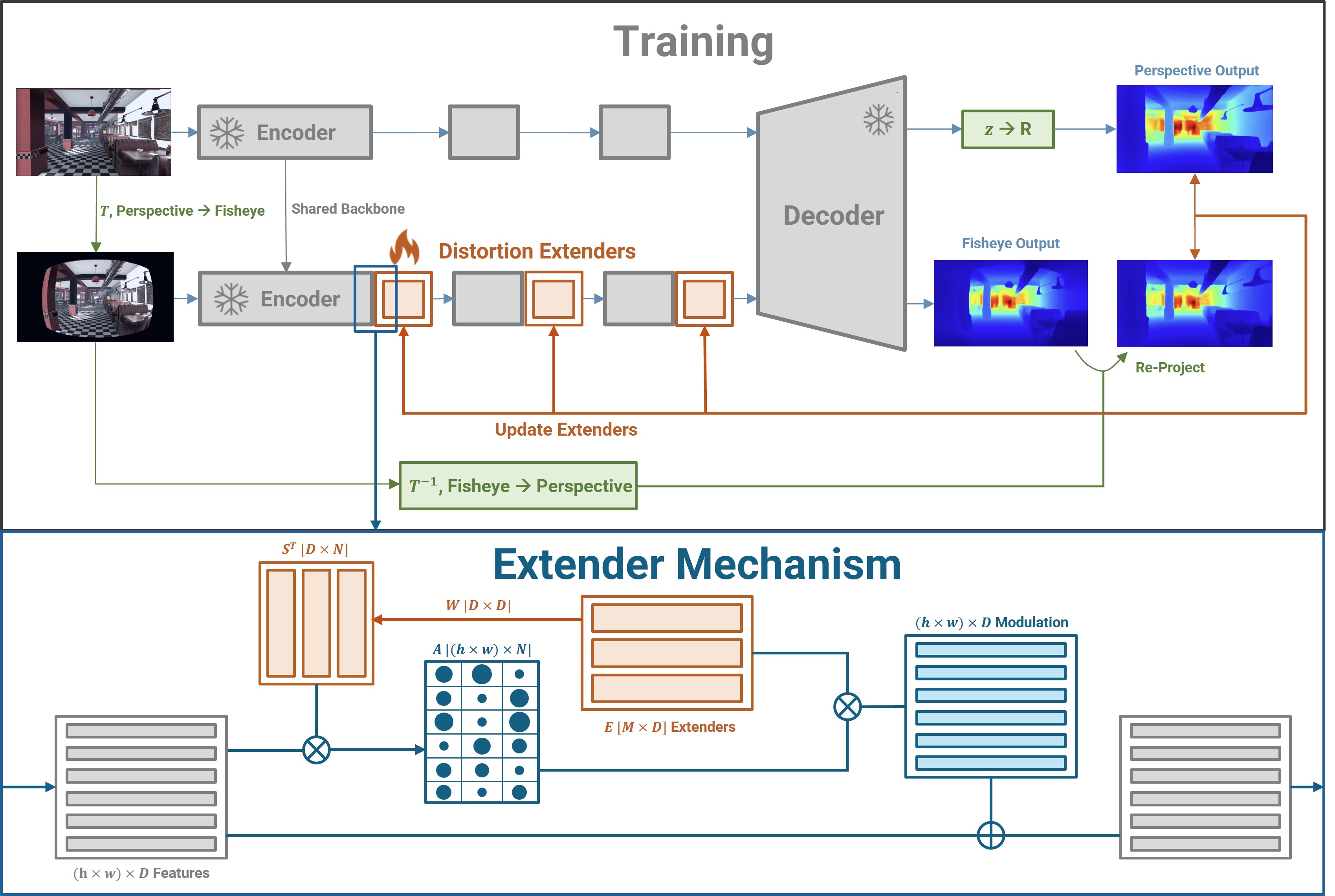}
  \caption{\textbf{Distortion Extenders.} The figure illustrates the DEX training pipeline and architecture. During training, we pass perspective inputs and fisheye inputs augmented by DEX into a frozen backbone model. The fisheye features select convex combinations of Extenders for translation in the latent space. We optimize the Distortion Extenders to minimize the difference between the fisheye and perspective output.}
  \label{fig:methoda}
\end{figure*}

\textbf{Formalization.} We assume that we are given a model pretrained on perspective images, where its architecture comprises an image encoder $f(\cdot)$ and a task decoder $g(\cdot)$. We note that $f$ and $g$ are parameterized functions, but for ease of notation, we omit their parameters as both sets are fixed or frozen in our method. Let $I : \Omega \subset \mathbb{R}^2 \rightarrow \mathbb{R}^3$ denote an input RGB image, where $\Omega$ denotes the image space. Inference for a task is given by $\hat{y} = g(f(I))$ for the estimate $\hat{y} \in \mathbb{R}^{H \times W \times C}$, where $H$ and $W$ are its spatial dimensions. In the case of regression tasks, e.g., depth estimation, $C=1$ and $\hat{y}$ denotes a 2.5D range or depth map. In the case of classification tasks, e.g., open-vocabulary segmentation, $C$ varies depending on the number of input classes or text labels and $\hat{y}$ are the logits corresponding to affinities between the image and text features. 

To extend the pretrained model to handle fisheye images with similar fidelity as perspective images, we propose \textit{Distortion Extenders} (DEX) to model distortions observed in the image and to correct for the covariate shift in values of the latent embeddings stemming from fisheye distortion. To this end, the crux of our method lies in a set of modulators that translate the latent embeddings of fisheye images such that they resemble those of perspective images; the result of which is a latent alignment of fisheye image embeddings to the distribution of perspective image embeddings. This naturally necessitates two steps: (1) determine the degree of distortion and their effect on the latent embeddings and (2) determine the modulation necessary to correct for their effect. To make the modeling of the broad range of distortions tractable, each Extender serves as a discrete ``center'' in this continuous space, where convex combinations of multiple Extenders determine the translation necessary to correct for the effect of image distortion on latent embeddings of a fisheye image. As such, DEX is comprised of two primary components: (1) a ``soft'' selection mechanism that is parameterized by a  weight matrix $\mathbf{W} \in \mathbb{R}^{D \times D}$ and (2) a set of \(M\) Extender modulation parameters $\mathbf{E} \in \mathbb{R}^{M \times D}$, where $D$ denotes the dimensions of latent embeddings encoded by $f$. Thus, the full architecture of DEX is \(\theta = \{\mathbf{E}^{(l)}, \mathbf{W}^{(l)}\}_{l=1}^L\) for \(L\) layers. An overview of our method is shown in Fig.~\ref{fig:methoda}.

\textbf{Extender Mechanism.} 
Let \(\mathbf{X}^{(l)} = f^{(l)}(\mathbf{X}^{(l-1)})\) be the output of the \(l\)th layer or block of an image encoder (where \(\mathbf{X}^{(0)} = {I}\)). We learn Distortion Extenders \(\theta\), to modulate the layer's output, \(\mathbf{X}^{(l)} \in \mathbb{R}^{(h \times w) \times D}\), for all \(L\) layers. To reduce over-parametrization, we use lower rank matrices to represent \(\mathbf{W} = \mathbf{W}_1^\top \mathbf{W}_2\), where \(\mathbf{W}_1, \mathbf{W}_2 \in \mathbb{R}^{d \times D}\) with \(d << D\). As the covariate shift stems from changes in calibration, we hypothesize a low dimensional matrix can encode the transformation. Our modulation after each transformer block reads:
\begin{equation}
\mathbf{S} = \mathbf{E} \mathbf{W},\quad
\mathbf{A} = softmax(\frac{\mathbf{X} \mathbf{S}^\top}{\sqrt{D}}), \ \ \ \mathbf{X} \leftarrow \mathbf{X} + \mathbf{AE},
\end{equation}
which allows us to add a convex combination of Extenders to every latent embedding (whether token or feature vector) in \(\mathbf{X}\), weighted by the affinity \(\mathbf{A}\) between each latent embedding and each Extender's selection in \(\mathbf{S}\). The modulated embeddings are then used in the next layer and any skip connections:
{\small
\vspace{-3mm}
\begin{equation}
\mathbf{X}^{(l+1)}
  = f^{(l+1)}(\mathbf{X}^{(l)} + (\mathbf{A}\mathbf{E})^{(l)})
  = f^{(l+1)}\!\big(f^{(l)}(\mathbf{X}^{(l-1)} + (\mathbf{A}\mathbf{E})^{(l-1)}) + (\mathbf{A}\mathbf{E})^{(l)}\big)
\end{equation}
}

Distortion Extenders can be thought of as "centers" that model the latent calibration differences between perspective and fisheye cameras. As we take convex combinations of Extenders, they form a boundary around a range of possible modulations to correct for an image's distortion in the latent space. The intuition is that an input image's features are able to select the weights of the combination of Extenders that represents its own distortion. We study this further in Sec.~\ref{sec:analysis}, where we use this principle to decode Distortion Extenders into KB distortion coefficients using an image's selected weights.

\textbf{Training.}
We aim to optimize our learnable Distortion Extenders, \(\theta\), at each of the \(L\) encoder blocks to align fisheye image latent embeddings with those of perspective images. We use a self-supervised training scheme, based on AugUndo \cite{wu2024augundo}. We start with calibrated, perspective images \(I\), and distort them with a fisheye transformation \(\mathbf{T}\) to yield a synthetic fisheye image counterpart \(I_f=\mathbf{T} \circ I\). We freeze a backbone model \(f\) trained on perspective images. Then, we retrieve the output of the perspective image \(\hat{y} = g(f(I))\), and the output of the fisheye image augmented with DEX, e.g., \(\hat{y}_f = g(f_{\theta}(I_f))\). We then apply the inverse fisheye transformation $\mathbf{T}^{-1}$ on $\hat{y}_f$ in the output space to facilitate comparison between the perspective output and the spatially aligned fisheye output in the perspective reference frame. We formulate the optimization as:
\begin{equation}
\argmin_{\theta}
\frac{1}{N}\sum_{n=1}^{N}\mathcal{L}(\hat{y}^{(n)}, \mathbf{T}^{-1} \circ \hat{y}_f^{(n)}).
\label{eq:loss_generic}
\end{equation}
Distortion Extenders can be applied to both reconstruction and recognition tasks. Thus, we derive two different forms of the learning objective from Eq.~\ref{eq:loss_generic}.

\textbf{Monocular Depth Estimation.} In the case of depth estimation, our Distortion Extenders also learn to re-project the output from a Cartesian coordinate system to a spherical one.  We treat our perspective output as the supervision signal \(\hat{z} = g(f(I))\), and retrieve the output of the fisheye image with Distortion Extenders \(\hat{R}_f = g(f_{\theta}(I_f))\). However, the baseline model's output is in depth. Thus, we convert the perspective output from \(z\)- to \(R\)-coordinates using the following process for each point on the depth map:
\begin{equation}
\mathbf{p'}=\begin{bmatrix}x' & y' & 1\end{bmatrix}^\top,\quad
\mathbf{P_{proj}}=\mathbf{K}^{-1} \mathbf{p'} = \begin{bmatrix}X/z & Y/z & 1\end{bmatrix}^\top,
\end{equation}
\begin{equation}
R=|z \cdot \mathbf{P_{proj}}|=|\mathbf{P}| = \sqrt{X^2+Y^2+z^2}.
\end{equation}
Finally, we use the following loss to optimize the Distortion Extenders \(\theta\):
\begin{equation}
\mathcal{L}(\hat{R}^{(n)}, \mathbf{T} \circ \hat{R}_f^{(n)})=\log(|\hat{R}^{(n)} - \mathbf{T} \circ \hat{R}_f^{(n)}|+1).
\end{equation}

\textbf{Open-Vocabulary Segmentation.} Most architectures for open-vocabulary segmentation involve separately encoding the image \(f_i(I)\) and a set of text labels \(f_l(t)\), representing the desired classes for classification. Then, some form of similarity is computed between the image embeddings and the language embeddings in order to assign a class to each pixel of the image. Thus, instead of computing our loss on the entire network's output, which includes a multiplication with the language modality, we compare the perspective and fisheye image embeddings:
\begin{equation}
\mathcal{L}(f_i^{(L)}(I^{(n)}), \mathbf{T}^{-1} \circ  f^{(L)}_{i,\theta}(I_f^{(n)})) = \log(|f_i^{(L)}(I^{(n)})-\mathbf{T}^{-1} \circ  f^{(L)}_{i,\theta}(I_f^{(n)}))|+1).
\end{equation}

\textbf{Inference.} At inference time, no rectification or knowledge of calibration is necessary to handle fisheye images. For depth estimation, we simply attach our Distortion Extenders to the backbone model and produce our output reconstruction \(\hat{R} = f_{\theta}(I_f)\). The same follows for segmentation to produce our image embedding \(f_{i, \theta}(I_f)\), which is later compared with language embeddings \(f_l(t)\) in the baseline model's architecture using the inner product. The additional overhead in storage and time for inference is minimal, as shown in Tab.~\ref{tab:computation}.

\section{Experiments}

\textbf{Datasets.} During the training process, we synthesize fisheye images using the KB distortion model, and optimize our Distortion Extenders by comparing with the perspective image output. As such, only calibrated, perspective images are necessary for training. Ground truth or real fisheye images are not necessary. We evaluate our method on real indoor and outdoor fisheye images, where we train just one set of Distortion Extenders for both domains, showing the ability of our method to generalize across 3D scenes. 

\underline{Training} datasets: \emph{NYUv2}~\cite{silberman2012indoor} provides a wide range of indoor perspective imagery; \emph{VOID}~\cite{wong2020unsupervised} includes perspective office, classroom, and stairwell scenes; \emph{IRS}~\cite{wang2021irs} contains synthetic, rendered perspective views of homes, restaurants, and retail environments; \emph{Hypersim}~\cite{roberts2021hypersim} offers high-quality photorealistic indoor scenes, including both residential and commercial, captured in a perspective projection. The \emph{Waymo} dataset~\cite{sun2020scalability} supplements these with a broad collection of perspective urban driving scenes.
\underline{Validation} datasets: \emph{ScanNet++}~\cite{yeshwanth2023scannet++} provides real fisheye images and ground-truth depth, collected via laser scanning and DSLR fisheye capture, enabling evaluation on authentic indoor fisheye data. \emph{KITTI-360}~\cite{liao2022kitti} features suburban driving scenes recorded with a multi-sensor rig that includes fisheye cameras, along with 3D ground-truth points for assessing reconstruction accuracy. \emph{WoodScape}~\cite{yogamani2019woodscape} offers similar driving scenes with 2D semantic instance labels. We use this dataset to evaluate performance in the open-vocabulary segmentation task. See Supp. Mat. for more details on datasets.

\textbf{Models.} Distortion Extenders are architecture- and task-agnostic. As such, we evaluate them with state-of-the-art monocular depth estimation models using the Vision Transformer architecture: MiDaS \cite{ranftl2019towards}, DepthAnything \cite{yang2024depthanything}, and UniDepthV2 \cite{piccinelli2025unidepthv2}. We also evaluate our Extenders with LSeg \cite{li2022language}, a state-of-the-art model for open-vocabulary semantic segmentation. To show that DEX is architecture agnostic, we additionally evaluate with CNN-based models, VNL \cite{yin2019enforcing} for monocular depth estimation and SED \cite{xie2024sed} for open-vocabulary segmentation. 

\begin{figure*}[t]
  \centering
  \includegraphics[width=\textwidth]{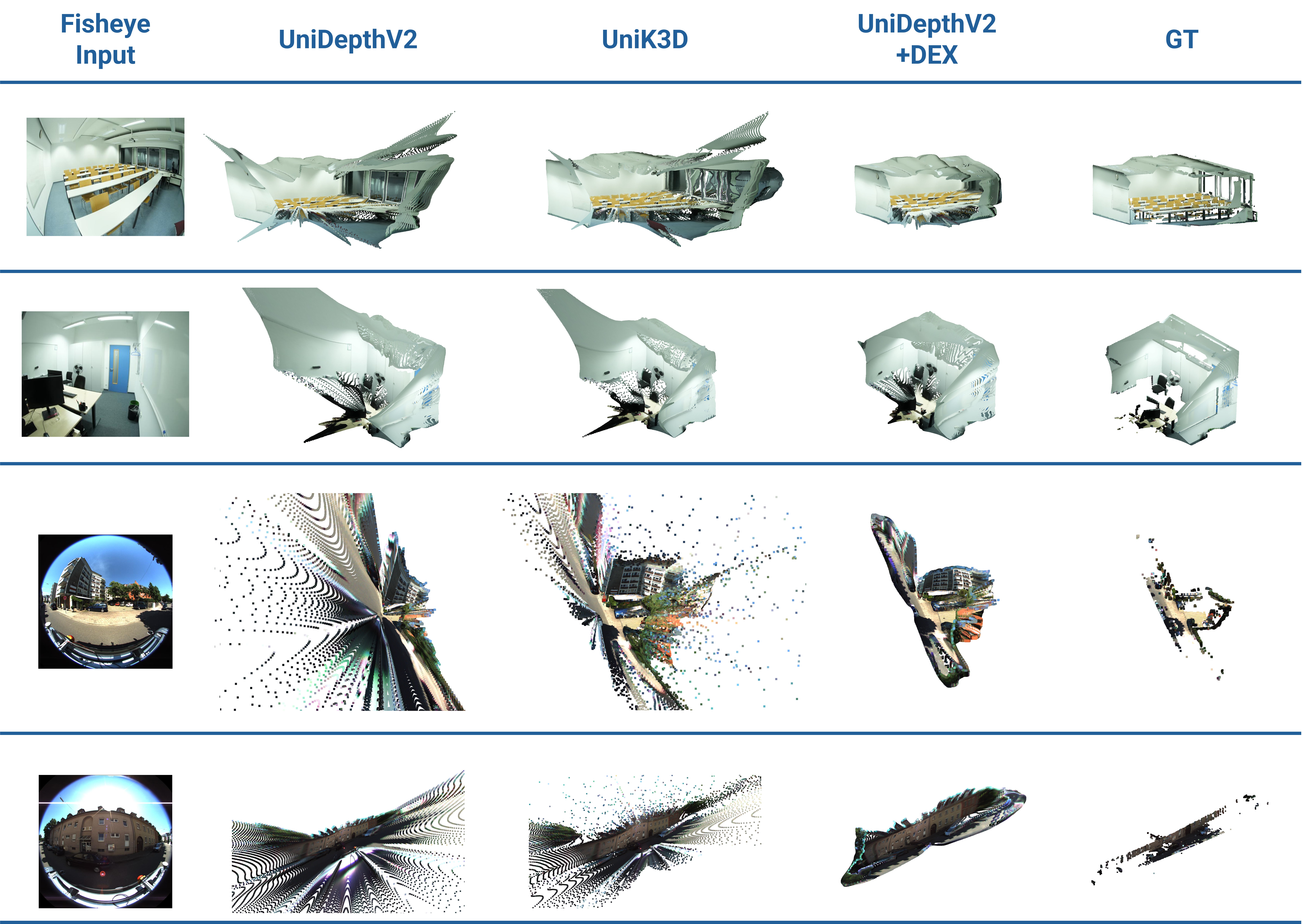}
  \caption{\textbf{MDE Qualitative Results.} We show 3D reconstructions of indoor and outdoor samples from the evaluation in Tab.~\ref{tab:main}. Previous methods struggle with the outermost regions in the fisheye images, which have the most severe distortion. DEX is more robust to these distortions. In the outdoor scenes with the highest distortions, we are still able to accurately reconstruct many of the points at the edges of the images.}
  \vspace{-0.2cm}
  \label{fig:qual}
\end{figure*}

\subsection{Evaluation on Fisheye Datasets}

We evaluate our method's performance when used with various MDEs in Tab.~\ref{tab:main}, testing on real fisheye images in both indoor and outdoor settings with the same set of Extenders. We also compare our method to UniK3D \cite{piccinelli2025unik3d}, a state-of-the-art method for fisheye monocular depth estimation. Additionally, we compare our method to LoRA adaptation and Calibration Tokens adapted for $R$. As there are currently no other models to perform segmentation specifically for fisheye images, we evaluate our method in conjunction with LSeg \cite{li2022language} and SED \cite{xie2024sed}, which perform open-vocabulary semantic segmentation. Note: We further adapted Calibration Tokens \cite{gangopadhyay2025extending} for LSeg for an additional comparison, but could not for VNL and SED as Calibration Tokens require Transformers.

We additionally compare DEX against a parameter-efficient LoRA adaptation of UniK3D to our training set. UniK3D already includes large-scale fisheye training data from ASE, aiMotive, HOI4D, and DL3DV, totaling approximately 3.263M examples, as well as ScanNet++ pinhole data. DEX achieves lower RMSE and higher $\delta_1$ on both ScanNet++ and KITTI-360. We hypothesize that UniK3D performance has largely saturated and that further adaptation with our mix of 200K examples makes little difference.

\begin{figure}[t]
    \centering
    \includegraphics[width=0.70\linewidth]{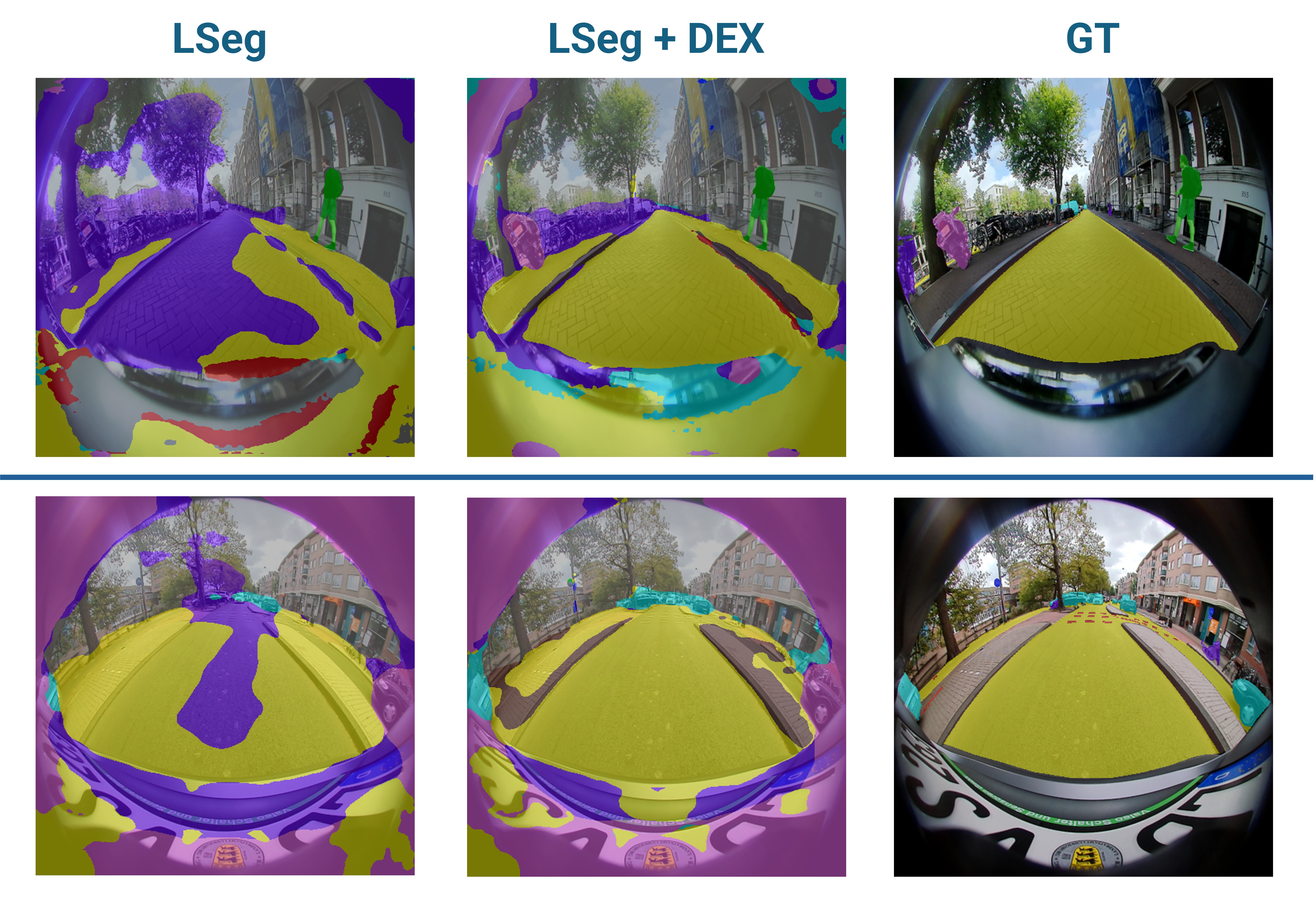}
    \vspace{-2.5mm}
    \caption{\textbf{Segmentation Qualitative Results.} We evaluate LSeg and SED with and without DEX on the WoodScape outdoor dataset. We see improvements in the open-vocabulary segmentation, highlighting DEX's generalizability to this recognition-based task. We evaluate on the valid regions of the images specified by the ground truth.}
    \label{fig:qual3}
    \vspace{-0.3cm}
\end{figure}

\textbf{Indoor MDE.} We evaluate the performance of MDE models with and without DEX on various scenes from  ScanNet++ \cite{yeshwanth2023scannet++}, an indoor fisheye dataset. We compare the output 3D reconstructions against the ground truth in the ground truth's Cartesian coordinate system. Across the baseline monocular depth estimators, DEX improves by 19\% in RMSE and 15\% in \(\delta_1\) on average over LoRA and Calibration Tokens. Additionally, when applied to UniDepthV2, our method improves over UniK3D by 8\% averaged over the metrics.

\begin{table*}[ht]
    \centering
    \caption{\textbf{MDE Performance on Indoor and Outdoor.} After training a set of Distortion Extenders, we test their zero-shot performance on ScanNet++ and KITTI-360, which contain real fisheye images with varying scenes and distortion parameters.}
    \label{tab:main}
    \vspace{-2mm}
    \scriptsize
    \begin{tabular}{c l l c c c}
        \toprule
        Testset & Experiment & Model & Train Dataset
        & RMSE $\downarrow$ & $\delta_1$ $\uparrow$ \\
        \midrule
        \multirow{17}{2.2mm}{\centering\rotatebox{90}{ScanNet++ \cite{yeshwanth2023scannet++}}}
        & Base Model & MiDaS \cite{ranftl2019towards} & Mix 1.4M 
         &0.672 &0.479  \\
         & \textbf{w/ LoRA} \cite{hu2022lora} & MiDaS \cite{ranftl2019towards} &  Mix 200K 
          & 0.561  & 0.599  \\
         & \textbf{w/ Calibration Tokens} \cite{gangopadhyay2025extending} & MiDaS \cite{ranftl2019towards} &  Mix 200K 
          & 0.413  & 0.688  \\
          & \textbf{w/ Distortion Extenders} & MiDaS \cite{ranftl2019towards} &  Mix 200K 
          & 0.397  & 0.702  \\  
        & Base Model &DepthAnything \cite{yang2024depthanything} & Mix 63.5M 
         & 0.705   & 0.454  \\
         & \textbf{w/ LoRA} \cite{hu2022lora} &DepthAnything \cite{yang2024depthanything} &  Mix 200K 
        & 0.706   & 0.445  \\
         & \textbf{w/ Calibration Tokens} \cite{gangopadhyay2025extending} &DepthAnything \cite{yang2024depthanything} &  Mix 200K 
        & 0.479   & 0.667  \\
        & \textbf{w/ Distortion Extenders} &DepthAnything \cite{yang2024depthanything} &  Mix 200K 
        & 0.414 & 0.708  \\
         & Base Model &UniDepthV2 \cite{piccinelli2025unidepthv2} & Mix 16M 
         & 0.329   & 0.671  \\
         & \textbf{w/ LoRA} \cite{hu2022lora} &UniDepthV2 \cite{piccinelli2025unidepthv2} &  Mix 200K   
         &{0.235}  &{0.854}  \\
         & \textbf{w/ Calibration Tokens} \cite{gangopadhyay2025extending} &UniDepthV2 \cite{piccinelli2025unidepthv2} &  Mix 200K   
         &{0.223}  &{0.841}  \\
         & \textbf{w/ Distortion Extenders} &UniDepthV2 \cite{piccinelli2025unidepthv2} &  Mix 200K   
         &\textbf{0.200}  &\textbf{0.872}  \\
         & Base Model &VNL \cite{yin2019enforcing} & NYUD-V2 29K 
         & 0.680   & 0.521  \\
         & \textbf{w/ Distortion Extenders} &VNL \cite{yin2019enforcing} &  Mix 80K   
         &0.372  &0.704  \\
         \cmidrule(lr){2-6}
         & {Comparison} & DepthAnyCamera \cite{guo2025depth}& Indoor 670K 
         &0.390   &0.852  \\
         & {Comparison} & UniK3D \cite{piccinelli2025unik3d}& Mix 12M   
         &0.223   &0.825  \\
         & \textbf{w/ LoRA} & UniK3D \cite{piccinelli2025unik3d}& Mix 200K   
         &0.218 &0.835  \\
        \midrule
        \multirow{17}{2.2mm}{\centering\rotatebox{90}{KITTI-360 \cite{liao2022kitti}}}
        &Base Model &MiDaS \cite{ranftl2019towards} & Mix 1.4M 
          & 6.111   & 0.312 \\
          & \textbf{w/ LoRA} \cite{hu2022lora} &MiDaS\cite{ranftl2019towards} &  Mix 200K   
          & 2.479  &0.601  \\
         & \textbf{w/ Calibration Tokens} \cite{gangopadhyay2025extending} &MiDaS\cite{ranftl2019towards} &  Mix 200K   
          & 2.588  & 0.699  \\
          & \textbf{w/ Distortion Extenders} &MiDaS\cite{ranftl2019towards} &  Mix 200K   
          & 2.451  &0.657  \\
         &Base Model &DepthAnything \cite{yang2024depthanything} & Mix 63.5M 
         & 6.484   & 0.318  \\
         & \textbf{w/ LoRA} \cite{hu2022lora} &DepthAnything \cite{yang2024depthanything} &  Mix 200K 
         & 2.897  &0.550  \\
         & \textbf{w/ Calibration Tokens} \cite{gangopadhyay2025extending} &DepthAnything \cite{yang2024depthanything} &  Mix 200K 
         & 2.224  & 0.676  \\
         & \textbf{w/ Distortion Extenders} &DepthAnything \cite{yang2024depthanything} &  Mix 200K 
         & 2.022  &0.745  \\
         &Base Model &UniDepthV2 \cite{piccinelli2025unidepthv2} & Mix 16M 
          & 7.093   & 0.262  \\
        & \textbf{w/ LoRA} \cite{hu2022lora} &UniDepthV2 \cite{piccinelli2025unidepthv2} &  Mix 200K  
         &{1.916} & {0.771} \\
         & \textbf{w/ Calibration Tokens} \cite{gangopadhyay2025extending} &UniDepthV2 \cite{piccinelli2025unidepthv2} &  Mix 200K  
         &{1.788} & {0.763} \\
         & \textbf{w/ Distortion Extenders} &UniDepthV2 \cite{piccinelli2025unidepthv2} &  Mix 200K  
         &\textbf{1.663} & \textbf{0.842} \\
         & Base Model &VNL \cite{yin2019enforcing} & KITTI 24K 
         & 7.719   & 0.245  \\
         & \textbf{w/ Distortion Extenders} &VNL \cite{yin2019enforcing} &  Mix 80K   
         &2.222  &0.605  \\
         \cmidrule(lr){2-6}          
         & {Comparison} & DepthAnyCamera \cite{guo2025depth}& Outdoor 130K 
         &3.641   &0.789  \\   
         & {Comparison} & UniK3D \cite{piccinelli2025unik3d}& Mix 12M   
          &2.969   &0.812  \\     
          & \textbf{w/ LoRA} & UniK3D \cite{piccinelli2025unik3d}& Mix 200K   
         &2.802 &0.818  \\
        \bottomrule
    \end{tabular}
    \vspace{-3mm}
\end{table*}

\begin{table*}[h!]
    \centering
    \caption{\textbf{Open Vocabulary Segmentation.} We evaluate the zero-shot performance of our method on the WoodScape dataset, which contains real fisheye images with varying scenes and distortion parameters.}
    \label{tab:seg}
    \vspace{-2mm}
    \scriptsize
    \setlength{\tabcolsep}{4pt}
    \begin{tabular}{c l l c c c}
        \toprule
        Testset & Experiment & Model & Train Dataset
        & mIoU $\uparrow$ & weighted IoU $\uparrow$ \\
        \midrule
        \multirow{5}{*}{{WoodScape \cite{yogamani2019woodscape}}}
        & Baseline & LSeg \cite{li2022language} & Mix 500K 
         &0.305 & 0.819  \\
         & \textbf{w/ Calibration Tokens} & LSeg \cite{li2022language} &  Mix 50K 
          & {0.321}  & {0.823}  \\
         & \textbf{w/ Distortion Extenders} & LSeg \cite{li2022language} &  Mix 50K 
          & \textbf{0.362}  & \textbf{0.838}  \\
        & Baseline & SED \cite{xie2024sed} & Mix 120K 
         &0.439 &0.829  \\
         & \textbf{w/ Distortion Extenders} & SED \cite{xie2024sed} &  Mix 50K 
          & \textbf{0.449}  & \textbf{0.832}  \\
        \bottomrule
    \end{tabular}
    \vspace{-1mm}
\end{table*}

\textbf{Outdoor MDE.} We also evaluate on the KITTI-360 outdoor fisheye image dataset. These images feature much higher FoV and as such, much more severe fisheye distortion. Predicting in spherical coordinates is especially helpful in this case. We compare all output 3D reconstructions against the ground truth in the ground truth's spherical coordinate system. On average, we see an 11\% improvement in RMSE and 11\% improvement in \(\delta_1\) over LoRA and Calibration Tokens across the baseline models. Additionally, the same set of Extenders is used for the indoor and outdoor setting, which demonstrates how DEX takes advantage of the existing backbone model's inherent generalization.

\textbf{Open-Vocabulary Segmentation.} We evaluate the LSeg and SED open-vocabulary semantic segmentation models on the WoodScape segmentation dataset with and without Distortion Extenders. We query the model with the following classes during evaluation: road, lane markings, curb, human, cyclist, cars, bicycle, motorcycle, street sign, other. We find that DEX improves over Calibration Tokens by around 13\% in the mIoU metric. The experiments are shown in Tab.~\ref{tab:seg}.

\subsection{DEX as Task-Agnostic}

To further evaluate whether DEX is task-agnostic, we apply it to PanopticDepth \cite{gao2022panopticdepth}, a multitask framework for depth estimation and depth-aware panoptic segmentation. We jointly optimize DEX by summing both depth and segmentation loss terms. The depth term similarly compares DEX's fisheye depth output to PanopticDepth's perspective depth output. The segmentation term is computed in this case by comparing DEX's fisheye semantic logits to PanopticDepth's perspective semantic logits. We are able to reuse the same training set for the experiment because we use the PanopticDepth baseline outputs as supervision. As shown in Tab.~\ref{tab:task_agnostic_panopticdepth}, DEX improves performance in both depth estimation on KITTI-360 and segmentation on WoodScape.

\begin{table}[!t]
\centering
\setlength{\tabcolsep}{6pt}
\renewcommand{\arraystretch}{0.80}
\caption{\textbf{DEX with PanopticDepth.} We apply DEX to PanopticDepth and evaluate both depth estimation and segmentation on WoodScape.}
\label{tab:task_agnostic_panopticdepth}
\begin{tabular}{lcccc}
\toprule
\multirow{2}{*}{Method}
& \multicolumn{2}{c}{Depth}
& \multicolumn{2}{c}{Segmentation} \\
\cmidrule(lr){2-3} \cmidrule(lr){4-5}
& RMSE $\downarrow$ & $\delta_1 \uparrow$
& mIoU $\uparrow$ & weighted IoU $\uparrow$ \\
\midrule
PanopticDepth \cite{gao2022panopticdepth}      & 7.273 & 0.267 & 0.160 & 0.300 \\
\textbf{w/ Distortion Extenders} & \textbf{3.583} & \textbf{0.354} & \textbf{0.282} & \textbf{0.833} \\
\bottomrule
\end{tabular}
\end{table}

\section{Analysis}
\label{sec:analysis}

\begin{figure*}[t]
  \centering

  \begin{minipage}[t!]{0.6\linewidth}
  \centering
  \captionof{table}{\textbf{Ablation Experiments.} We evaluate DEX's performance without low rank decomposition, and without extension to spherical coordinates.}
  \label{tab:ablation}
  \scriptsize
  \begin{tabular}{l l c c}
      \toprule
        Dataset &Method &RMSE &$\delta_1$ \\
        \midrule
        \multirow{3}{*}{ScanNet++}
        & Distortion Extenders & \textbf{0.200}  &\textbf{0.872}\\
         & w/o \(\hat{R}\) Extension & 0.249 & 0.792  \\
         & w/o Low-Rank Decomposition & 0.216  & 0.849 \\
         \midrule
        \multirow{3}{*}{KITTI-360} 
        & Distortion Extenders & \textbf{1.663}  &\textbf{0.842}\\
         & w/o \(\hat{R}\) Extension & 7.024 & 0.271 \\
         & w/o Low-Rank Decomposition & 1.777  & 0.800\\
      \bottomrule
  \end{tabular}
  \vspace{-3mm}
\end{minipage}\hfill
\begin{minipage}[th!]{0.35\linewidth}
  \centering
  \captionof{table}{\textbf{Computational Overhead.} Additional memory and inference time of Distortion Extenders over the backbone models. DEX adds minimal complexity.}
  \label{tab:computation}
  \small
  \scriptsize
  \setlength{\tabcolsep}{0pt}
  \begin{tabular}{lcc}
      \toprule
      Model & UniDepthV2 & LSeg \\
      \midrule
      \makecell[l]{Base memory}     & 0.7\,GB              & 2.9\,GB \\
      \makecell[l]{DEX memory}  & 2.8\,MB     & 6.1\,MB \\
      \makecell[l]{Inference}   & +0.3\,ms              & +2.21\,ms \\
      \bottomrule
  \end{tabular}
  \vspace{-3mm}
\end{minipage}

  \vspace{5mm} 
    \includegraphics[height=0.27\textheight]{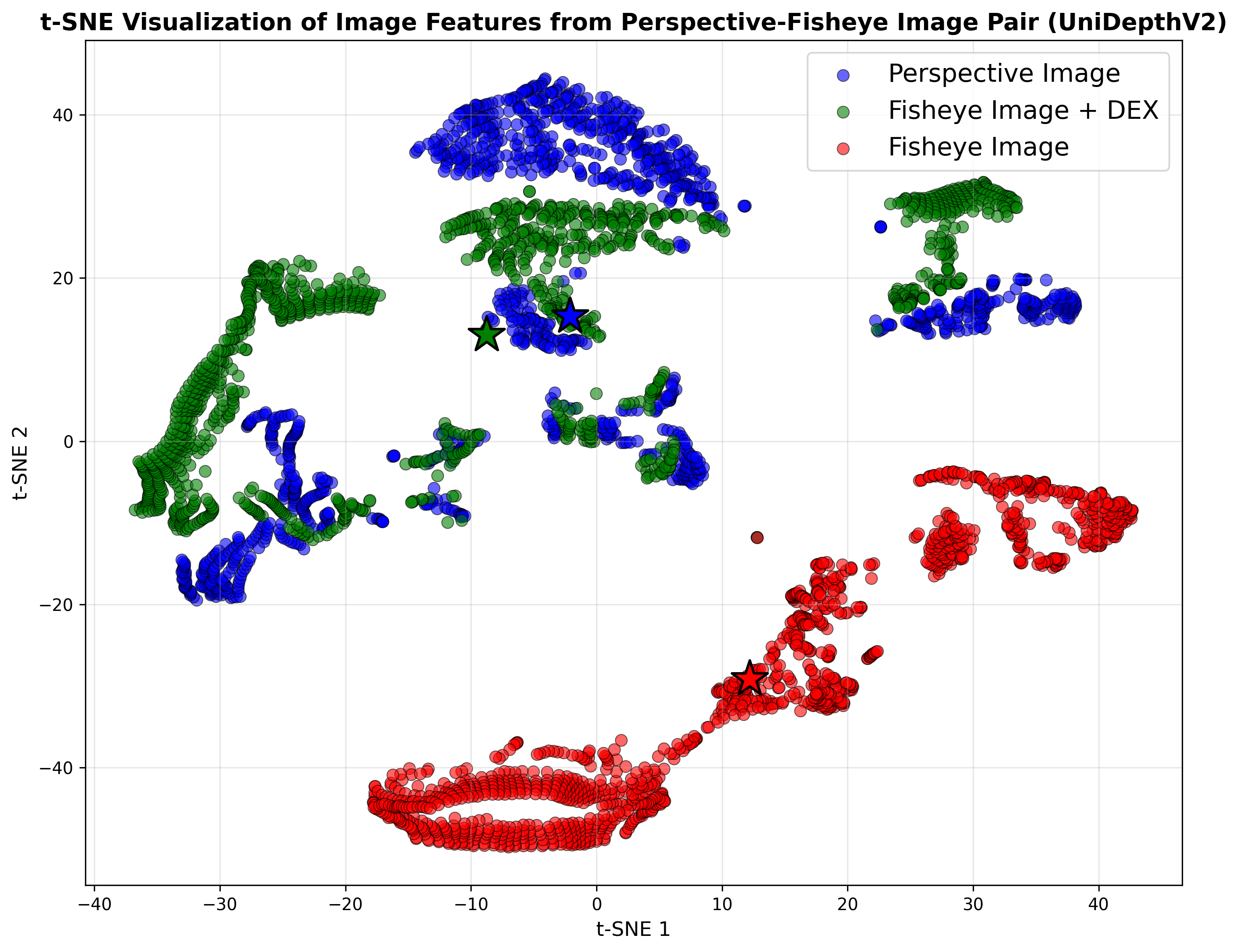}
    \vspace{-3mm}
    \caption{\textbf{t-SNE Plot of Fisheye and Perspective Image Features.} When using DEX, fisheye image features align much closer with corresponding perspective features.}
    \label{fig:tsne}
    \vspace{-4mm}

\end{figure*}

\textbf{Ablation Study.} In order to characterize the contributions of various architectural decisions, we ablate them and evaluate the performance of depth estimation with the UniDepthV2 model. As expected, learning to estimate depth in the Cartesian coordinate system has an especially negative effect in the outdoor setting, which has a much wider FoV with stronger fisheye distortion. Additionally, without the low-rank decomposition, there is a slight performance degradation due to over-parametrization. See Tab.~\ref{tab:ablation} for the full results.

\textbf{Feature Alignment.} We claim that DEX is able to make fisheye images more conducive to models trained on perspective imagery by modulating fisheye image features towards the distribution of perspective features. We verify this claim in Fig.~\ref{fig:tsne}, where we use a t-SNE dimensionality reduction to visualize the UniDepthV2 image features of a perspective image, and its corresponding fisheye image with and without DEX. When the model is augmented with DEX, the fisheye image features align very closely with the perspective image features.

\textbf{DEX Decoding.} To better understand the internal representations learned by DEX, we attempt to decode the Extenders into KB distortion coefficients for an image input. We posit that because DEX learns to align fisheye features to perspective features, Distortion Extenders implicitly encode KB distortion coefficients. Thus, we should be able to find the Extenders that encode a particular image's distortion coefficients using the weightings of its activated Extenders. 

As such, we pass an image input through UniDepthV2 with our trained DEX, and extract the affinities \(A\) along with the Extenders \(\mathbf{E}\) for all \(L\) layers. We then use a simple linear probe \(\mathbf{W}_D \in \mathbb{R}^{D \times 4}\) to decode each Extender into four values \(\begin{bmatrix} \alpha_1^{(i)} & \alpha_2^{(i)} & \alpha_3^{(i)} & \alpha_4^{(i)}\end{bmatrix}_{i=1}^{M} = \mathbf{E}\mathbf{W}_D\). For each layer, we take a linear combination across the \(\alpha\) values for each Extender, using the affinities \(A\) averaged across image features. We then predict the final KB distortion coefficients \(\begin{bmatrix} k_1 & k_2 & k_3 & k_4\end{bmatrix}\) with one more linear combination across layers, with a learnable weighting \(\mathbf{W}_L \in \mathbb{R}^L\). As we also optimize DEX to extend the baseline model to spherical coordinates, this learnable weighting across layers helps us select the layers with Extenders most relevant to the fisheye distortion correction.

We optimize \(\mathbf{W}_D\) and \(\mathbf{W}_L\) using the same training data in the DEX training pipeline, and evaluate on the same ScanNet++ evaluation set. As seen in Tab.~\ref{tab:decoding_results}, with a simple linear regressor, we are able to predict \(k_1\) with the highest accuracy, and \(k_4\) with lower accuracy. The degradation pattern from \(k_1\) to \(k_4\) aligns with the fisheye distortion effect from each parameter. Changes in \(k_1\) will result in a much greater visual effect than changes in \(k_4\), e.g., moderate changes in $k_4$ do not yield visual differences, so the earlier coefficients dominate the discriminativeness in determining distortion. As such, Distortion Extenders encode later distortion coefficients with wider uncertainty.

\begin{figure}[t]
    \centering
    \includegraphics[width=0.8\linewidth]{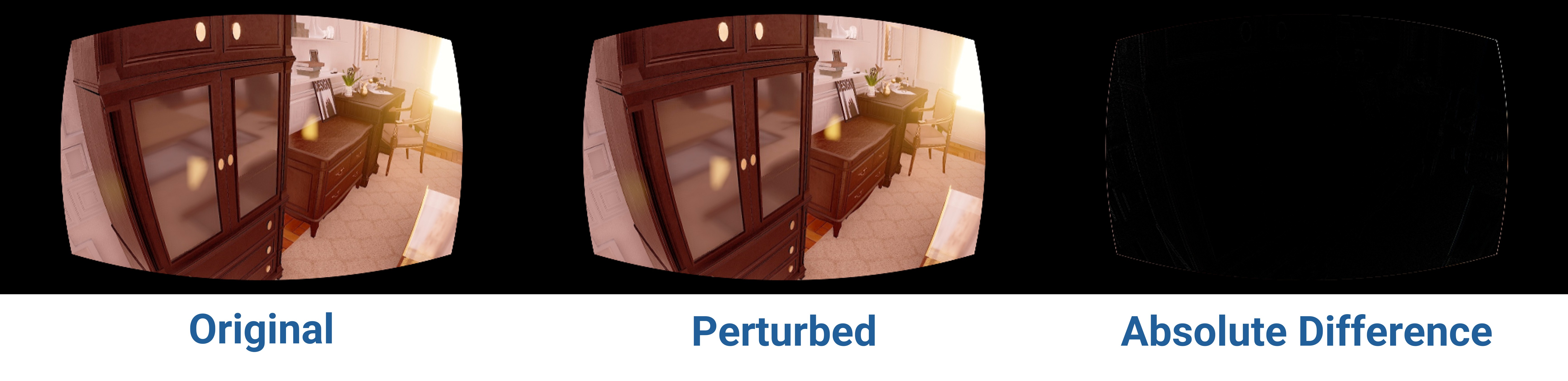}
    \caption{\textbf{Error in KB Distortion Prediction Visualized.} We provide an example for the visual difference in distortion when we generate a fisheye image with a set of distortion parameters representative of ScanNet++, and when we change all of the distortion coefficients by our average error in prediction. Absolute difference between the images generated by the two different sets of distortion coefficients has little visual difference. Absolute difference is best viewed in 5$\times$.}
    \label{fig:perturbed}
\end{figure}

We show the visual difference for perturbing the distortion coefficients in an image by the error percentages produced by our method in Fig.~\ref{fig:perturbed}. There is very little difference caused by this shift in distortion parameters, indicating difficulty in recovering them with higher fidelity.

\begin{table}[t]
\centering
\caption{\textbf{KB Distortion Parameter Decoding.} Distortion Extenders are able to recover KB distortion parameters for an image input.}
\setlength{\tabcolsep}{5pt}
\begin{tabular}{lccccc}
\toprule
ScanNet++ & $k_1$ & $k_2$ & $k_3$ & $k_4$ & Average \\
\midrule
MSE $\downarrow$       & $2.04{\times}10^{-6}$ & $1.53{\times}10^{-6}$ & $3.53{\times}10^{-7}$ & $8.87{\times}10^{-9}$ & $9.81{\times}10^{-7}$ \\
MAE $\downarrow$       & $1.17{\times}10^{-3}$ & $9.88{\times}10^{-4}$ & $5.27{\times}10^{-4}$ & $7.05{\times}10^{-5}$ & $6.88{\times}10^{-4}$ \\
MAPE (\%) $\downarrow$ & 3.91 & 14.30 & 41.86 & 78.61 & 34.67 \\
\bottomrule
\end{tabular}
\label{tab:decoding_results}
\end{table}

\section{Discussion}

DEX provides a simple but effective mechanism for adapting perspective-trained vision foundation models to fisheye imagery without modifying the backbone or performing explicit geometric rectification at test time. By operating directly in feature space, our approach mitigates the projection mismatch that arises from high-FoV lenses. The improvements observed across multiple backbones and datasets (Tab.~\ref{tab:main}, ~\ref{tab:seg}) suggest that many of the errors in distorted views stem from a systematic feature-space shift. More broadly, this work illustrates the flexibility of our training pipeline. The same self-supervised alignment framework of using a frozen backbone, synthetic domain shift, inverse geometric alignment, and lightweight extenders can be reused across architectures and tasks.

\textbf{Limitations.} During training, we do not need ground truth or fisheye images. However, this means that our performance is upper-bounded by the quality of the backbone models we use as our supervision signal, which implies we must carefully select training data that the backbone model can produce high-fidelity outputs for. Luckily, FMDEs and open-vocabulary segmentation models have been trained on tens to hundreds of millions of images. This allows us to casually select training datasets. We hypothesize that one must take care in dataset selection for smaller-scale models.

\section*{Acknowledgements} 
This work is supported by a Google Gift Award, NSF-2112562 Athena AI Institute and the Global Industrial Technology Cooperation Center (GITCC) through a grant agreement with the Korea Institute for Advancement of Technology (KIAT), project number P0028922.

\bibliographystyle{splncs04}
\bibliography{main,visionlab}

\newpage

\appendix

\begin{center}
    {\Large{\textbf{ 
        From Perspective to Fisheye Depth Estimation and Open-Vocabulary Segmentation  \\
        \vspace{0.5cm}
        SUPPLEMENTARY MATERIAL
    }}}
    \vspace{1cm}
\end{center}

\section{Affinity Maps of Distortion Extenders}

\begin{figure}
    \centering
    \includegraphics[width=0.8\linewidth]{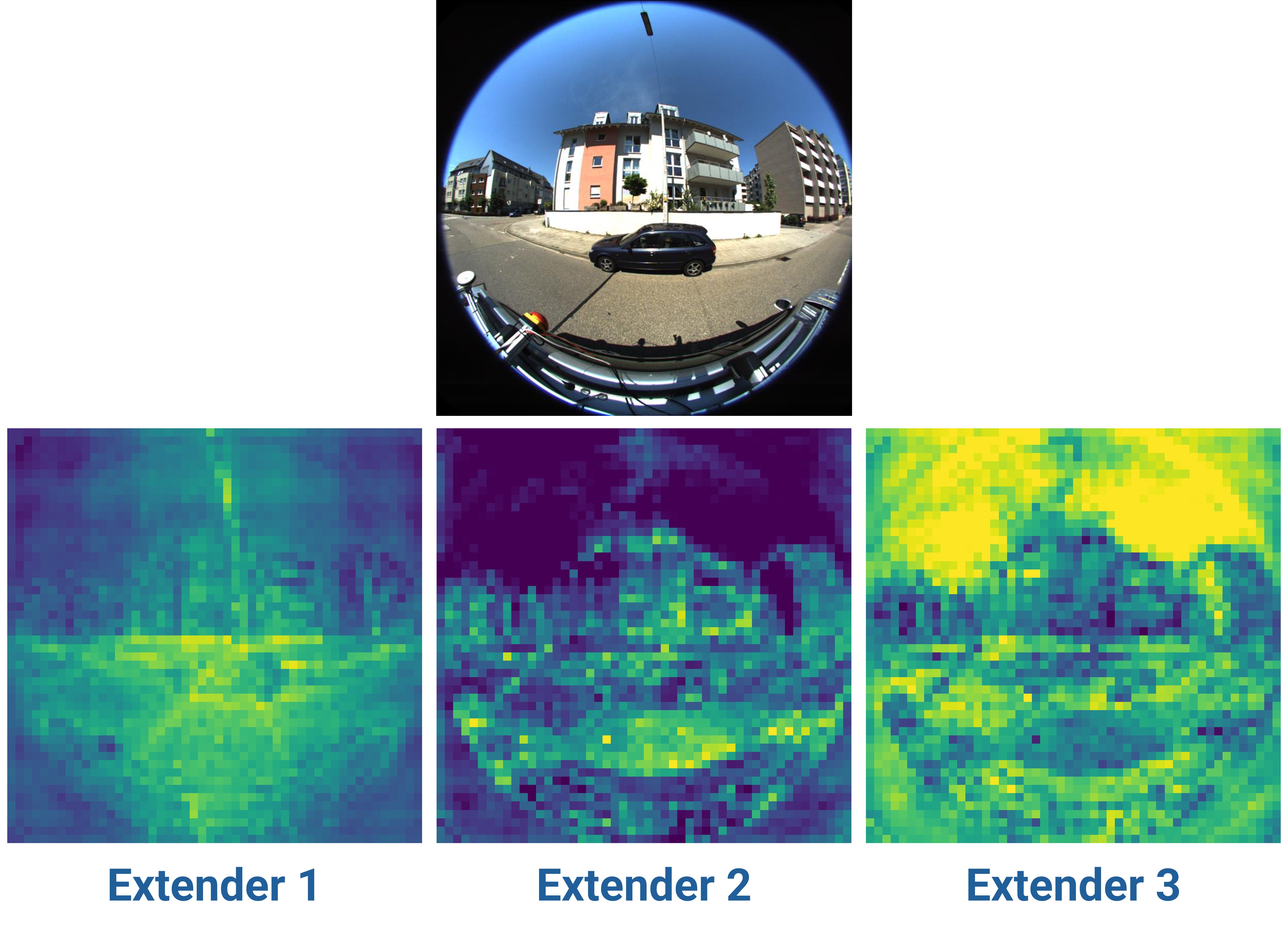}
    \caption{\textbf{Affinity Maps of DEX.} Each Extender exhibits different affinity toward different regions of the fisheye image features, enabling a selective, yet cooperative modulation of the fisheye features to align them back to those of perspective images.}
    \label{fig:affinity}
\end{figure}

Fig.~\ref{fig:affinity} visualizes the affinity of various Distortion Extenders toward patch embeddings of the input image. Different Extenders learn to modulate different parts of the image. 

\section{Sensitivity Studies}

In Tab.~\ref{tab:ablation_combined}, we study sensitivity to number of extenders $M$, to decomposition dimension $d$, and to using DEX at a subset of layers for UniDepthV2. For the layer placement strategies, we tested DEX at every other layer ("Every Other"), just the first layer ("First"), and a single trained set of Extenders shared and reused across all layers ("Shared"). We also study sensitivity to different sets of labels used in open-vocabulary segmentation evaluation for LSeg in Tab.~\ref{tab:woodscape_label_sensitivity}.

We note the study on different sets of labels (Tab.~\ref{tab:woodscape_label_sensitivity}) demonstrates that regardless of labels, DEX always improves performance. As stated in the Limitations section in Sec. 6 of the main paper and the extended discussion in Sec. \ref{sec:extended_discussion_limitations}, we note that the improvement of DEX is related to that of the base model (LSeg). As such, differences in performance across label sets are artifacts of the base model rather than evidence of inherent sensitivity in DEX.

\begin{table}[!t]
\centering
\setlength{\tabcolsep}{6pt}
\renewcommand{\arraystretch}{0.60}
\caption{\textbf{Sensitivity study on hyperparameters and layer placement.} We analyze the effect of the number of extenders $M$, decomposition dimension $d$, and different layer placement strategies on monocular depth estimation performance for ScanNet++ and KITTI-360. Within the "Layers" section of the table, "Every Other" refers to inserting DEX at every other encoder block; "First" refers to only using DEX at the first encoder block; "Shared" refers to training one set of Extenders and reusing it for every encoder block.}
\label{tab:ablation_combined}
\begin{tabular}{lcccccc}
\toprule
& \multirow{2}{*}{$M$} & \multirow{2}{*}{$d$}
& \multicolumn{2}{c}{ScanNet++}
& \multicolumn{2}{c}{KITTI-360} \\
\cmidrule(lr){4-5} \cmidrule(lr){6-7}
& & & RMSE $\downarrow$ & $\delta_1 \uparrow$
& RMSE $\downarrow$ & $\delta_1 \uparrow$ \\
\midrule
\multirow{9}{*}{{Sensitivity}}
& 12 & 64 & 0.209 & 0.845 & 1.704 & 0.826 \\
& 12 & 20 & 0.200 & 0.860 & 1.663 & 0.842 \\
& 12 & 10 & 0.207 & 0.858 & 1.695 & 0.844 \\
& 12 & 8  & 0.207 & 0.852 & 1.709 & 0.847 \\
& 12 & 4  & 0.212 & 0.847 & 1.746 & 0.840 \\
& 32 & 20 & 0.201 & 0.860 & 1.685 & 0.815 \\
& 16 & 20 & 0.203 & 0.859 & 1.723 & 0.814 \\
& 8  & 20 & 0.204 & 0.858 & 1.740 & 0.796 \\
& 4  & 20 & 0.218 & 0.832 & 1.708 & 0.818 \\
\addlinespace
\midrule
\addlinespace
\multirow{3}{*}{{Layers}}
& \multicolumn{2}{c}{Every Other}
& 0.207 & 0.847 & 1.653 & 0.833 \\
& \multicolumn{2}{c}{First}
& 0.250 & 0.792 & 1.979 & 0.681 \\
& \multicolumn{2}{c}{Shared}
& 0.233 & 0.812 & 1.665 & 0.779 \\
\addlinespace
\bottomrule
\end{tabular}
\end{table}

\begin{table}[!h]
\centering
\setlength{\tabcolsep}{5pt}
\renewcommand{\arraystretch}{1.0}
\caption{\textbf{Sensitivity to label wording on WoodScape open-vocabulary segmentation.} We evaluate how different textual label sets affect the performance of LSeg with DEX on WoodScape. We note that the sensitivity of outcomes for different label sets is more of a property of the base model (LSeg) than it is of DEX.}
\label{tab:woodscape_label_sensitivity}
\resizebox{\columnwidth}{!}{%
\scriptsize
\begin{tabular}{p{5.9cm}ccc}
\toprule
\textbf{Label Set} & \textbf{Method} & \textbf{mIoU $\uparrow$} & \textbf{weighted IoU $\uparrow$} \\
\midrule

\multirow{2}{=}{\texttt{[}miscellaneous, street, lanemarks, curbstone, person, rider, vehicles, bike, motorbike, traffic sign\texttt{]}}
& LSeg   & 0.289 & 0.812 \\
& w/ DEX & 0.305 & 0.822 \\[12pt]

\multirow{2}{=}{\texttt{[}background, roadway, road markings, sidewalk edge, pedestrian, riding person, automobiles, cycle, moto, road sign\texttt{]}}
& LSeg   & 0.227 & 0.792 \\
& w/ DEX & 0.293 & 0.816 \\[12pt]

\multirow{2}{=}{\texttt{[}other, road, lane markings, curb, human, cyclist, cars, bicycle, motorcycle, street sign\texttt{]}}
& LSeg   & 0.305 & 0.819 \\
& w/ DEX & {0.362} & {0.838} \\[2pt]

\bottomrule
\end{tabular}
}
\end{table}

\section{Implementation Details}

For the depth estimation experiments, we train 12 Distortion Extenders (\(M = 12\)) for each transformer block, and for VNL, we used 72, 18, 9, 4, and 2 Extenders for the 5 residual blocks in the encoder. For segmentation, we train 24 Distortion Extenders for each transformer block in LSeg \cite{li2022language}, and 24, 12, 6, and 3 Extenders for each resolution of CNN encoder blocks in SED \cite{xie2024sed}. For all experiments, we learn projection matrices \(
\mathbf{W_1}, \mathbf{W_2}\) with a smaller inner dimension \(d = 20\). We train using the Adam optimizer with a learning rate of \(10^{-3}\), and a batch size of 32 for depth estimation and 16 for segmentation. We create our synthetic fisheye training images by randomly sampling distortion coefficients from the following ranges: k1  from -1.3 to  -0.005, k2 from -1.3 to -0.005, k3 from -1.3 to -0.005, and k4 from -0.5 to 0.0, inclusive. We trained DEX for 8 epochs using 4 RTX3090 GPUs. Code and checkpoints for DEX will be released for reproducibility.

\section{Evaluation Protocols}

As mentioned in the main paper, all evaluations on depth estimation are conducted in the coordinate frame of the ground truth: ScanNet++ in \(z\)-coordinates, and KITTI-360 in \(R\)-coordinates (resolutions of 462 $\times$ 616 and 700 $\times$ 700, respectively). Hence, we convert our depth estimate from \(R\) to \(z\) for ScanNet++, but no conversion is needed for KITTI-360. Similarly, outputs of FMDEs (natively in \(z\)) are converted for KITTI-360, but no change is needed for ScanNet++. The \(z\)- to \(R\)-coordinate transformation for evaluation follows Eqs. 4 and 5 in the main paper, but we must undo fisheye distortion after inverting the intrinsics. We formulate this conversion and the conversion from \(R\) to \(z\) below.

\subsection{R to z Transformation}

For each pixel \(\mathbf{p'}\) in the \(R\) fisheye distance map, we start by recovering the originally projected point by inverting the intrinsics \(\mathbf{K}\) and the fisheye distortion model transformation \(\mathbf{T}\) (KB \cite{kannala2004generic} for the case of ScanNet++):

\begin{equation}
\mathbf{p'}=\begin{bmatrix}x'\\y'\\1\end{bmatrix},\quad
\mathbf{P_{dist}}=\mathbf{K}^{-1} \mathbf{p'} = \begin{bmatrix}u\\v\\1\end{bmatrix},
\end{equation}

\begin{equation}
\mathbf{P_{proj}} = \mathbf{T}^{-1} \mathbf{P_{dist}} = \begin{bmatrix}X/z\\Y/z\\1\end{bmatrix}
\end{equation}

Now, \(\mathbf{P_{proj}}\) is in the direction of the actual 3D point \(\mathbf{P}\), so we back-project by scaling with \(R\) and extract the \(z\) component to get the depth of the pixel:

\begin{equation}
\mathbf{P} = \frac{R}{|\mathbf{P_{proj}}|} \cdot \mathbf{P_{proj}} = \begin{bmatrix}X\\Y\\z\end{bmatrix}.
\end{equation}

\subsection{z to R Transformation}

For this transformation, we start with the above process to recover \(\mathbf{P_{proj}}\) (with the MEI \cite{mei2007single} model for KITTI-360). We then scale by $z$ and take the magnitude, similarly to the training process:

\begin{equation}
R=|z \cdot \mathbf{P_{proj}}|=|\mathbf{P}| = \sqrt{X^2+Y^2+z^2}.
\end{equation}

\section{Dataset Details}

\noindent\textit{Training Datasets:}

\textbf{NYUv2}~\cite{silberman2012indoor} contains 464 indoor scenes spanning homes, offices, and shared spaces, producing roughly 400k RGB–depth pairs at 640\(\times\)480 resolution. NYUv2 is a standard indoor benchmark and forms one of our primary supervised training sources.

\textbf{IRS}~\cite{wang2021irs} offers a large suite of synthetic indoor environments, from compact apartments to larger commercial layouts. Up to 103{,}316 rendered frames with accurate depth provide complementary coverage that broadens scene diversity beyond real indoor scans.

\textbf{VOID}~\cite{wong2020unsupervised} (Visual Odometry and Indoor Depth) supplies around 58k pairs of RGB image and depth maps captured in corridors, classrooms, and open common areas.

\textbf{Hypersim}~\cite{roberts2021hypersim} is a physically based synthetic dataset comprising approximately 77k photorealistic RGB–depth pairs. The scenes include carefully modeled geometry and lighting across diverse architectural settings, enabling the model to observe highly controlled yet visually rich interiors before evaluation on real data.

\textbf{Waymo}~\cite{sun2020scalability} contributes about 230k frames with synchronized camera and LiDAR across varied outdoor driving conditions. Although originally targeted at autonomous driving perception tasks, we include Waymo to broaden the distribution of our training data to wide-range outdoor imagery featuring complex lighting and large depth ranges.

\noindent\textit{\\Testing Datasets:}
Our evaluation focuses on real-world datasets featuring fisheye or wide-FOV imaging. 

\textbf{ScanNet++}~\cite{yeshwanth2023scannet++} extends the ScanNet family with additional indoor scenes and fisheye acquisitions. Its fisheye depth ground truth allows us to directly stress-test our method under strong lens distortion in cluttered interiors.

\textbf{KITTI-360}~\cite{liao2022kitti} is an outdoor, large-scale autonomous driving dataset equipped with panoramic fisheye cameras and high-resolution LiDAR. Its suburban and semi-rural sequences help measure how well our approach generalizes to wide-FOV outdoor imagery in challenging real driving scenarios.

\textbf{WoodScape}~\cite{yogamani2019woodscape} is a multi-camera fisheye dataset specifically built for automotive perception, providing four surround-view fisheye cameras with full 360° coverage as well as LiDAR, IMU, and GNSS measurements. It includes annotations for nine tasks—such as semantic segmentation, depth estimation, 2D/3D object detection, motion segmentation, soiling detection, and visual odometry/SLAM—with over 10{,}000 images annotated at the instance level and more than 100{,}000 frames annotated for other tasks.

\noindent\textit{\\Training sets:}

\textbf{Mix 200k for Monocular Depth Estimation} incorporates images from all of the training datasets. We take subsamples from each dataset in the training set (subsample 25k from NYUv2, 60k from IRS, 30k from VOID, 60k from Hypersim, and 25k from Waymo). Note: we follow the subsampling strategy of Calibration Tokens \cite{gangopadhyay2025extending} for fair comparison.

\textbf{Mix 80k for Monocular Depth Estimation} removes IRS and Hypersim because VNL was not trained on synthetic data. Including synthetic data introduces a sim-to-real gap, which degrades accuracy of estimates during training. This reflects the stated limitation of our self-supervised training scheme as discussed in "Limitations" in the Sec. 6 ("Discussion") of the main paper.

\textbf{Mix 50k for Open-Vocabulary Segmentation} removes IRS, Hypersim, and VOID to speed up training time due to computational resource limitations.

\begin{figure*}[!t]
  \centering
  \includegraphics[width=0.7\textwidth]{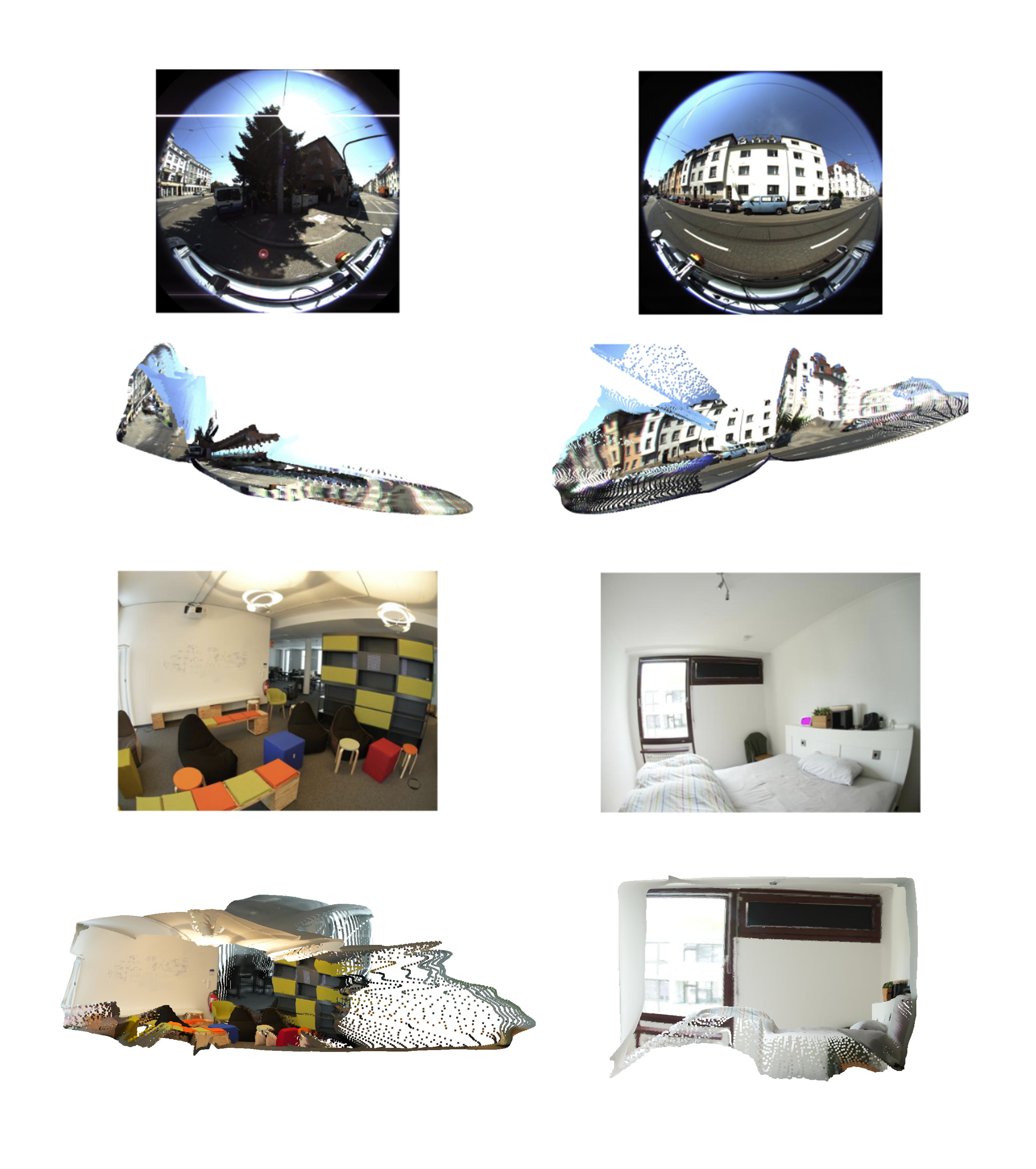}
  \caption{\textbf{Additional Qualitative Examples.} We show additional 3D reconstructions of diverse indoor and outdoor samples using DEX.}
  \label{fig:supp_qual}
\end{figure*}

\section{Additional Qualitative Examples}
We provide additional qualitative figures for both indoor (ScanNet++) and outdoor (KITTI-360) scenarios in Fig. \ref{fig:supp_qual}. All samples are created using the UniDepthV2 model with DEX. Our method is able to recover the L-shaped road (top left) as well as the straight road (top right) for outdoors. We are also able to get the farther portion of the room (bottom left) and straight edges in the room (bottom right) for indoors.

\begin{table}[!t]
\centering
\caption{
        \textbf{Error metrics for depth estimation and open-vocabulary segmentation.} These evaluation metrics compute the error between predicted output values and the ground truth. For depth estimation metrics (e.g. RMSE and $\delta_1$), $\hat{d}$ denotes the prediction and $d$ the ground truth. For open-vocabulary segmentation metrics (e.g., mIoU, weighted IoU), TP denotes true positive, FP denotes false positive, FN denotes false negative, and $n_k$ denotes the sum of pixels in the ground truth for the class $k$.}

\resizebox{0.6\columnwidth}{!}{%
\begin{tabular}{l l}
\toprule
\textbf{Metric} & \textbf{Definition} \\
\midrule
RMSE\,$\downarrow$ &
\(\displaystyle 
\sqrt{\frac{1}{|\Omega|} \sum_{p \in \Omega} 
\bigl(\hat{d}(p) - d(p)\bigr)^2}
\)
\\[0.6em]
\(\delta_1\)\,$\uparrow$ &
\(\displaystyle 
\frac{1}{|\Omega|} \sum_{p \in \Omega} 
\mathbf{1}\!\Bigl(
\max\!\Bigl(\frac{\hat{d}(p)}{d(p)}, 
\, \frac{d(p)}{\hat{d}(p)}\Bigr) 
< 1.25
\Bigr)
\) 
\\
$\text{mIoU}\uparrow$ & $\displaystyle \frac{1}{K} \sum_{k=1}^{K}
\frac{TP_k}{TP_k + FP_k + FN_k}$
\\
$\text{weighted IoU}\uparrow$ &  $\displaystyle 
\sum_{k=1}^{K} w_k \,
\frac{TP_k}{TP_k + FP_k + FN_k},$ 
$w_k = \frac{n_k}{\sum_{j=1}^{K} n_j}$
\\
\midrule
\end{tabular}%
}
\label{tab:depth_metrics}
\end{table}

\section{Evaluation Metrics}
Table~\ref{tab:depth_metrics} provides the mathematical definitions for evaluation metrics on monocular depth estimation and open-vocabulary segmentation.
For monocular depth estimation, we adopt standard benchmarking metrics to evaluate our predictions. Specifically, we use \emph{Root Mean Squared Error} (RMSE), which aggregates the magnitude of errors in linear space. Lower values in this metric indicate better accuracy. Additionally, we measure a threshold-based accuracy \(\delta_1\), which quantifies the percentage of pixels whose predicted output values lie within a bound of the ground-truth values. For open-vocabulary segmentation, we adopt mean intersection-over-union (mIoU) and weighted intersection-over-union (weighted IoU). mIoU computes the average intersection-over-union across all classes, treating each class equally regardless of its frequency. Weighted IoU accounts for class imbalance by weighting each class according to its ground-truth proportion.

\section{Comparison to Standard Undistortion Pipeline}

We compare DEX to a standard undistortion preprocessing pipeline, followed by inference with UniDepthV2 in Tab.~\ref{tab:undistortion_comparison}. Since undistortion changes the effective field of view and can reduce scene coverage, we alter our evaluation protocol to use Chamfer distance and F1 score to measure both accuracy and completeness of the reconstructed point cloud. DEX substantially improves over using undistortion as preprocessing on both ScanNet++ and KITTI-360.

\begin{table}[!h]
\centering
\setlength{\tabcolsep}{6pt}
\renewcommand{\arraystretch}{0.80}
\caption{\textbf{Comparison to standard undistortion preprocessing pipelines.} We compare undistortion preprocessing followed by inference with UniDepthV2 against UniDepthV2 with DEX using Chamfer distance and F1 score.}
\label{tab:undistortion_comparison}
\begin{tabular}{llccc}
\toprule
Base Model & Method & Dataset & Chamfer $\downarrow$ & F1 $\uparrow$ \\
\midrule
UniDepthV2 & Undistortion & ScanNet++ & 0.360 & 0.317 \\
UniDepthV2 & DEX          & ScanNet++ & \textbf{0.117} & \textbf{0.841} \\
\midrule
UniDepthV2 & Undistortion & KITTI-360 & 2.428 & 0.196 \\
UniDepthV2 & DEX          & KITTI-360 & \textbf{0.343} & \textbf{0.887} \\
\bottomrule
\end{tabular}
\end{table}

\section{Extended Discussion on Limitations and Future Work}
\label{sec:extended_discussion_limitations}
In the main paper, we mentioned that a limitation of our method is that it is bounded by the performance of the baseline model. We illustrate this point with the VNL model, which has separate checkpoints trained exclusively for indoors and outdoors. We evaluate the VNL model (NYUv2 and KITTI) with DEX on both the indoor and outdoor testing sets in Tab.~\ref{tab:limits}. As expected, we found that the performance of indoors-trained VNL with DEX degrades on outdoor data, and vice versa. This is because the models are trained on specific domains; hence, using it to generate self-supervision for a domain it was not trained on creates an out-of-distribution setting with respect to the training dataset (either NYUv2 \cite{silberman2012indoor} or KITTI \cite{geiger2013vision}). This degrades the supervision signal, leading to less improvement for the domain that the model was not originally trained on.

\begin{table}[h]
\centering
\setlength{\tabcolsep}{6pt}
\renewcommand{\arraystretch}{0.60}
\caption{\textbf{Limitations.} We evaluate the performance of the indoor and outdoor VNL models with DEX on both the indoor and outdoor test sets.}
\label{tab:limits}
\begin{tabular}{ccccc}
\toprule
\multirow{2}{*}{VNL} & \multicolumn{2}{c}{ScanNet++}
& \multicolumn{2}{c}{KITTI-360} \\
\cmidrule(lr){2-3} \cmidrule(lr){4-5}
& RMSE $\downarrow$ & $\delta_1 \uparrow$
& RMSE $\downarrow$ & $\delta_1 \uparrow$ \\
\midrule
NYUv2 & 0.372 & 0.704 & 6.283 & 0.301 \\[2pt]
KITTI & 1.153 & 0.377 & 2.222 & 0.605 \\
\bottomrule
\end{tabular}
\end{table}

The key insight behind the development of Distortion Extenders lies in the fact that existing pretrained foundation models \cite{li2022language,piccinelli2024unidepth,piccinelli2025unidepthv2,yang2024depthanything,yang2024depthanythingv2,xie2024sed} can already infer properties of the 3D scene. This begs the question of whether such models can be extended to a different projection model with the same (high) fidelity. As the forward image formation model can be seen as a composition of extrinsic (camera within the 3D scene) and intrinsic (camera internal parameters) factors, we see this as motivation to leverage existing knowledge of the model about (extrinsic) 3D scenes to extend them to (intrinsic) new cameras. Related are continual learning methods in cameras \cite{gangopadhyay2025extending,duan2026fisheye3r}, scenes \cite{rim2025protodepth}, and sensors \cite{rim2026radar,yang2024binding,tu2026unitac}. Because our method is self-supervised and task-agnostic, we also foresee it being applicable to test-time adaptation, e.g., for reconstruction \cite{park2024test,chung2025eta}, \cite{wang2021tent,zhang2025progressive}, where the parameters of extenders are learned online. Finally, our self-supervised loss can also be made unsupervised if given a sequence of images \cite{fei2019geo,garg2016unsupervised,godard2017unsupervised,godard2019digging,wong2019bilateral} and additional sensors \cite{liu2022monitored,park2026orcas, wong2020unsupervised,wong2021adaptive,wong2021learning,wong2021unsupervised,wu2024augundo}, making this approach suitable for deployment to platforms with heterogeneous sensors.

\end{document}